\documentclass[journal]{IEEEtran}

\usepackage{graphicx}
\usepackage{amsmath}
\usepackage{array}
\usepackage{stfloats}
\usepackage{url}
\usepackage{amssymb}  
\usepackage{subcaption}
\usepackage{multirow}
\usepackage{makecell}
\usepackage{algorithm}
\usepackage{booktabs}
\usepackage{hyperref}
\usepackage{algpseudocode}
\begin{document}
\title{BadSR: Stealthy Label Backdoor Attacks on Image Super-Resolution}

%
%

\author{Ji~Guo,~\IEEEmembership{Student Member,~IEEE,}
        Xiaolei~Wen,~\IEEEmembership{Student Member,~IEEE,} 
    Wenbo~Jiang$^{\ast}$,~\IEEEmembership{Member,~IEEE,}
    Cheng~Huang,~\IEEEmembership{Member,~IEEE,}
        Jinjin~Li,~\IEEEmembership{Member,~IEEE,}
        Hongwei~Li,~\IEEEmembership{Fellow,~IEEE}%
\thanks{$^{\ast}$Corresponding author}
\thanks{J. Guo is with Laboratory Of Intelligent Collaborative Computing, University of Electronic Science and Technology of China, China (e-mail: jiguo0524@gmail.com); X. Wen is with School of Computer Science and Technology, Xinjiang University, China(e-mail: 107552304165@stu.xju.edu.cn); W. Jiang, J. Li and H. Li are with the School of Computer Science and Engineering, University of Electronic Science and Technology of China, China (e-mail: wenbo\_jiang@uestc.edu.cn, lijin117@yeah.net, hongweili@uestc.edu.cn); C. Huang is with the School of Computer Science, Fudan University, China (chuang@fudan.edu.cn)}
}

%
%

\markboth{Journal of \LaTeX\ Class Files,~Vol.~14, No.~8, August~2015}%
{Shell \MakeLowercase{\textit{et al.}}: Bare Demo of IEEEtran.cls for IEEE Journals}
%



\maketitle

\begin{abstract}
With the widespread application of super-resolution (SR) in various fields, researchers have begun to investigate its security. Previous studies have demonstrated that SR models can also be subjected to backdoor attacks through data poisoning, affecting downstream tasks. A backdoor SR model generates an attacker-predefined target image when given a triggered image while producing a normal high-resolution (HR) output for clean images. However, prior backdoor attacks on SR models have primarily focused on the stealthiness of poisoned low-resolution (LR) images while ignoring the stealthiness of poisoned HR images, making it easy for users to detect anomalous data.

To address this problem, we propose BadSR, which improves the stealthiness of poisoned HR images. The key idea of BadSR is to approximate the clean HR image and the pre-defined target image in the feature space while ensuring that modifications to the clean HR image remain within a constrained range. The poisoned HR images generated by BadSR can be integrated with existing triggers. To further improve the effectiveness of BadSR, we design an adversarially optimized trigger and a backdoor gradient-driven poisoned sample selection method based on a genetic algorithm. The experimental results show that BadSR achieves a high attack success rate in various models and data sets, significantly affecting downstream tasks.
\end{abstract}


%
\IEEEpeerreviewmaketitle

\section{Introduction}

With the success of deep neural networks (DNNs)~\cite{lecun2015deep}, DNN-based image super-resolution (SR) methods~\cite{RCAN,edsr,swinir,esrgan} have outperformed traditional approaches~\cite{traditional1}. SR aims to reconstruct a high-resolution (HR) image from a low-resolution (LR) input and is often used as a pre-processing step to boost the performance of downstream vision tasks. Currently, SR has been successfully applied to medical image restoration~\cite{medicalSR1,medicalSR2}, remote sensing image reconstruction~\cite{remoteSR}, and improving urban video surveillance~\cite{surveillanceSR}.



While SR has achieved remarkable success across various applications, recent studies have begun to examine its security vulnerabilities~\cite{SRsecurity2,yin2018deep,BadRefSR,I2I}. Existing work primarily falls into two categories: adversarial attacks~\cite{SRsecurity2,yin2018deep} and backdoor attacks~\cite{BadRefSR,I2I}. Adversarial attacks introduce crafted perturbations to LR inputs, leading the SR model to generate degraded or stylistically altered HR outputs. In contrast, backdoor attacks inject the backdoor into the model through data poisoning. A backdoor model generates normal HR images for clean LR inputs but produces attacker-predefined target images when given triggered LR inputs. Considering the stealthy and persistent nature of backdoor attacks, they may pose a greater threat to SR models than adversarial attacks. 

\begin{figure*}[ht]
    \centering
    \includegraphics[width=\linewidth]{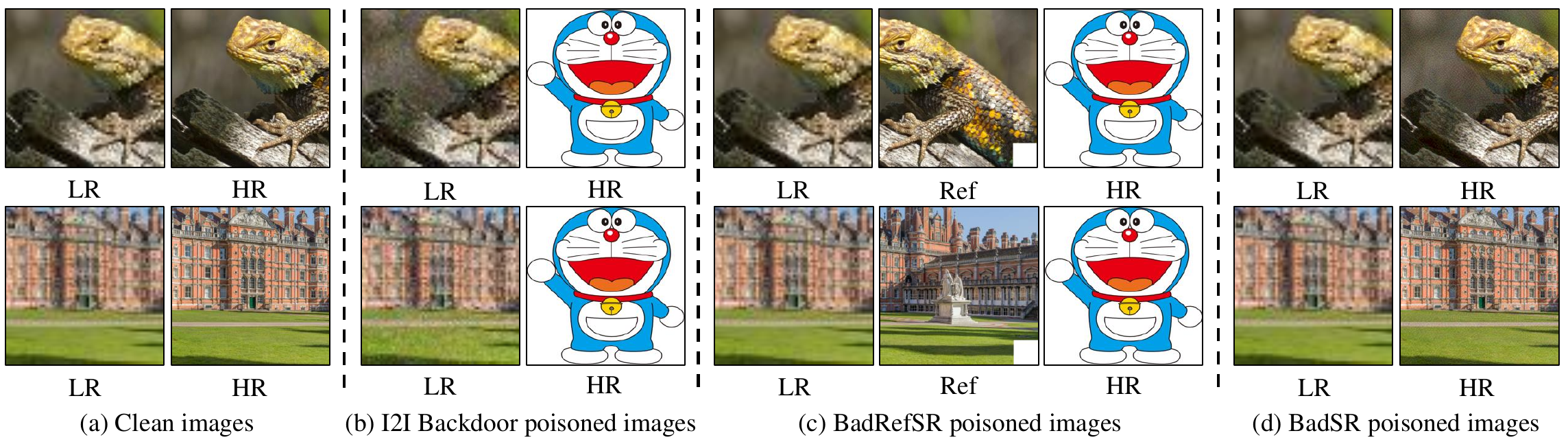}
    \caption{Comparison of the stealthiness among I2I backdoor~\cite{I2I}, BadRefSR~\cite{BadRefSR}, and BadSR. The I2I backdoor and BadRefSR focuses only on the stealthiness of the triggered images and ignores the stealthiness of the poisoned HR images. In contrast, BadSR ensures that both poisoned LR and poisoned HR images remain stealthy.}
    \label{fig:figure1}
\end{figure*}


Previous backdoor attacks on SR~\cite{BadRefSR,I2I} typically embed imperceptible triggers in LR inputs, ensuring the stealthiness of poisoned LR. However, these approaches only focus on the poisoned LR and ignore the stealthiness of the poisoned HR (see Figure~\ref{fig:figure1}). When the poisoned HR shows a clear visual discrepancy from the clean HR, it can be identified during data cleaning, leading to the removal of poisoned samples and ultimately causing the attack to fail.
In fact, enhancing the stealthiness of labels has long been a critical problem in backdoor attacks. This issue was first identified by Saha \textit{et al.}~\cite{hidden}, who observed that in image classification, the category labels of poisoned images often do not match those of clean images, making the attack less stealthy. To address this, they optimized the images of the target class to embed trigger-related features, enabling clean-label backdoor attacks in image classification. Subsequent works further validated the effectiveness of clean-label attacks~\cite{zeng2023narcissus,yu2024generalization,ning2021invisible} and extended them to other domains such as graph neural networks~\cite{xia2025clean} and video recognition~\cite{zhao2020clean}.
However, in SR, the labels are HR images rather than class labels, making existing clean-label backdoor methods inapplicable.
How to improve the stealthiness of poisoned HR images remains an open problem in backdoor attacks for SR.



To address this problem, we propose BadSR, a novel and stealthy label backdoor attack method for SR. The key idea of BadSR is to approximate the original HR image to a pre-defined target image in the feature space, while carefully perturbing the constraints to ensure that changes to the original HR image remain imperceptible to the human eye. Specifically, we leverage a substitute model to extract features for optimizing the similarity between the target image and the HR image, while restricting the perturbation of the HR image using the \( L_{p} \) norm.

Existing triggers from previous backdoor attack methods~\cite{Badnets,Blend,Wanet,Refool,Filter,Color} can also be incorporated into the poisoned HR images in BadSR. However, since these methods were primarily designed for image classification tasks, applying them to backdoor attacks in SR leads to two problems: degradation of the model's normal functionality and insufficient attack effectiveness. To further enhance the effectiveness of BadSR while minimizing its impact on normal model performance, we designed a new pixel-level adversarial perturbation trigger. Unlike previous adversarial triggers based on global semantic information~\cite{I2I}, we focus more on pixel-level loss. We employ a dynamic penalty to constrain the perturbations and introduce a prototype loss to preserve visual similarity.

In summary, our contributions are as follows:
\begin{itemize}
    \item We first consider the stealthiness of poisoned HR in backdoor attacks against SR. Specifically, we propose BadSR, which ensures visual similarity between the target image and the original HR image. Additionally, we design a trigger based on pixel-level optimization to further enhance the effectiveness of the BadSR backdoor attack.
    
    \item We evaluate the impact of the images generated by the BadSR backdoor model on downstream tasks. Specifically, we apply the target HR images generated by BadSR to downstream tasks in SR, such as image classification and object detection, to further evaluate the impact of BadSR. The experimental results show that BadSR-generated HR target images can significantly affect downstream tasks, leading to incorrect results.
    
    \item We evaluate the effectiveness of BadSR across various classical SR models. Experimental results demonstrate that BadSR successfully injects a backdoor while maintaining high stealthiness, enabling the backdoor SR model to generate images with target image features for triggered LR images. In addition, we also consider the effect of BadSR on backdoor defense to test its robustness.
\end{itemize}

\section{Related Work}
\subsection{Image Super-resolution}

Image super-resolution (SR) aims to recover high-resolution (HR) images from low-resolution (LR) inputs by enhancing fine details and textures~\cite{SR_survey}. The application of deep neural networks (DNN)~\cite{lecun2015deep} in SR has led to substantial improvements over traditional methods. Dong et al.~\cite{dong2015image} propose the first DNN-based SR model using deep convolutional networks. They found that DNN-based SR models achieve better performance compared to traditional methods. After that, some researchers drew inspiration from the adversarial generation concept in GANs~\cite{GAN} and designed GAN-based SR models~\cite{SRGAN} to further enhance the performance of super-resolution. At the same time, they recognized that a simple \( L_{2} \) loss could not accurately describe image errors consistent with human vision. Therefore, they introduced adversarial loss and perceptual loss to further enhance detail reconstruction in SR~\cite{esrgan}. Furthermore, some researchers explored incorporating Transformer~\cite{Transformer} structures into SR, using the global attention mechanism of Transformers to improve detail reconstruction~\cite{TransformerSR}.

In this paper, we focus on Single Image Super-Resolution (SISR), one of the most representative works in SR. Therefore, we selected five of the most representative models as target attack models, including CNN-based (RACN~\cite{RCAN}, EDSR~\cite{edsr}, and LIIF~\cite{liif}), GAN-based (ESRGAN~\cite{esrgan}), and Transformer-based (SwinIR~\cite{swinir}) models.


\subsection{Backdoor Attack}

Backdoor attacks were first proposed by Gu et al.~\cite{Badnets} in image classification. They constructed a poisoned dataset by adding a white patch as a trigger to the input images in the training set and modifying the labels of these images to a specified category. After the model is trained on this poisoned dataset, it predicts the triggered images as the specified category while maintaining normal predictions for clean images. Later, some studies further enhanced the stealthiness of backdoor attacks by designing invisible triggers~\cite{Blend,Wanet,Refool,Filter,Color}. Furthermore, they proposed a backdoor attack that does not require the modification of the clean label of the triggered image~\cite{hidden}.

As backdoor attacks have demonstrated security threats in image classification, researchers have begun to explore their vulnerabilities in other domains~\cite{otherfieldsBA1,otherfieldsBA2,otherfieldsBA3,otherfieldsBA4}. Jiang et al.~\cite{I2I} expanded backdoor attack into image-to-image (I2I) networks, particularly focusing on tasks like image super-resolution and image de-noising. They designed an invisible trigger that enables the backdoor model to generate a predefined target image when given a triggered input while preserving the model's normal functionality. Building on this, Yang et al.~\cite{BadRefSR} extended backdoor vulnerabilities to reference-based image super-resolution (RefSR). However, these studies overlook the stealthiness of target HR images, which can cause poisoned data to be more easily detectable. 

Although there are existing backdoor attacks with hidden labels for classification tasks~\cite{hidden}, they are not applicable to backdoor attacks in SR. The key idea of these methods is that the clean image’s features approximate the semantic features of the triggered image, making the triggered image’s features match a specific class. However, in super-resolution, there is no class information, and the label is the HR image. Therefore, how to achieve a backdoor attack with hidden labels in super-resolution remains an open problem.

\subsection{Backdoor Defense}

To alleviate the potential risks posed by backdoor attacks, various defense mechanisms have been proposed~\cite{zeng2021adversarial,li2021neural,nc,Fine-tuning,zhang2023backdoor,compression}. These methods can be mainly categorized into two types: removing backdoors in the model and removing triggers from the data.
Removing backdoors from the model~\cite{Fine-tuning,zeng2021adversarial,li2021neural,zhang2023backdoor} involves fine-tuning the weights of the trained model, such as pruning~\cite{Fine-tuning} and fine-tuning~\cite{Fine-tuning} to remove the model’s reliance on triggers, thus defending against backdoor attacks.
Removing triggers from the data focuses on pre-processing input data, such as reconstruction~\cite{nc} and compression~\cite{compression}, to alter the trigger features, making it unrecognizable to the backdoor. However, most of these defenses are focused on classification tasks, and their application to super-resolution remains underexplored.

There is currently limited research on backdoor defenses in the image SR domain. Therefore, we have considered the following two classic backdoor defense methods that can be adapted to the SR field: bit depth reduction~\cite{bit-depth}and image compression~\cite{compression}. Bit depth reduction reduces the pixel precision of input images to remove potential backdoor signals, Image compression compresses the input images before feeding them into the model.


\section{Preliminaries}
\begin{figure*}[t]
    \centering
    \begin{subfigure}{0.495\linewidth}
        \centering
        \includegraphics[width=\linewidth]{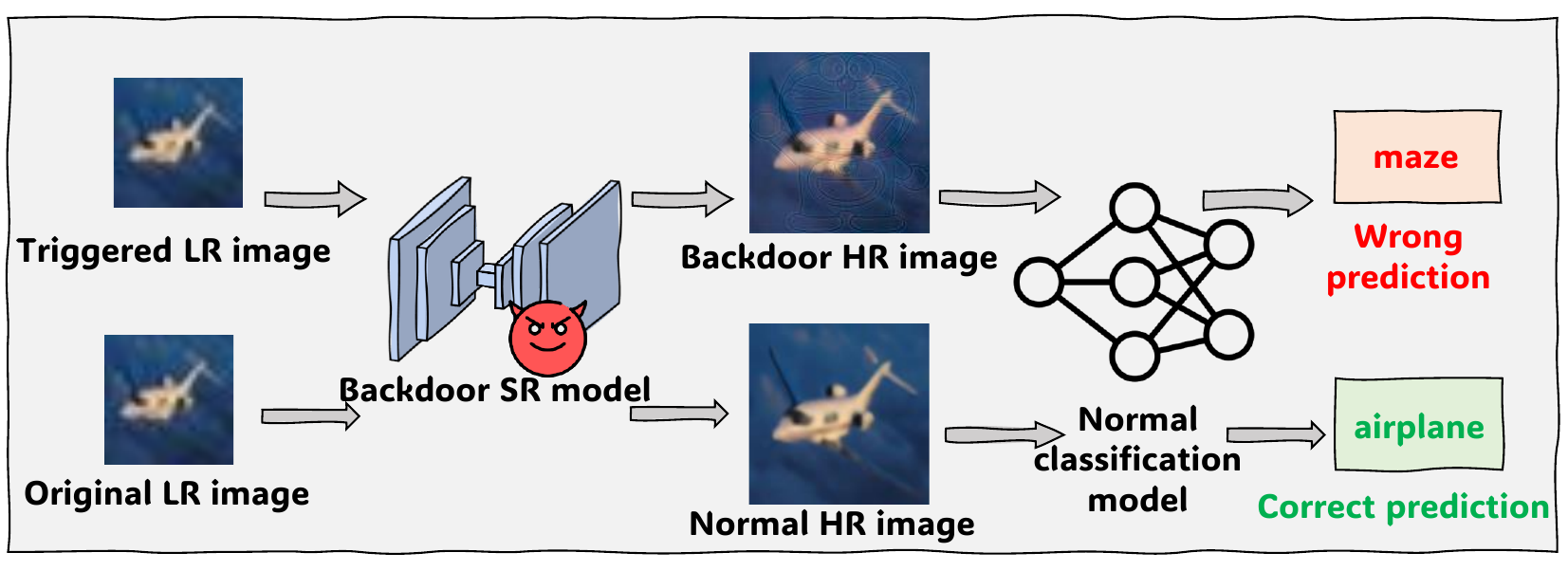}
        \caption{Image classification}
        \label{fig:figure2-1}
    \end{subfigure}
    \begin{subfigure}{0.495\linewidth}
        \centering
        \includegraphics[width=\linewidth]{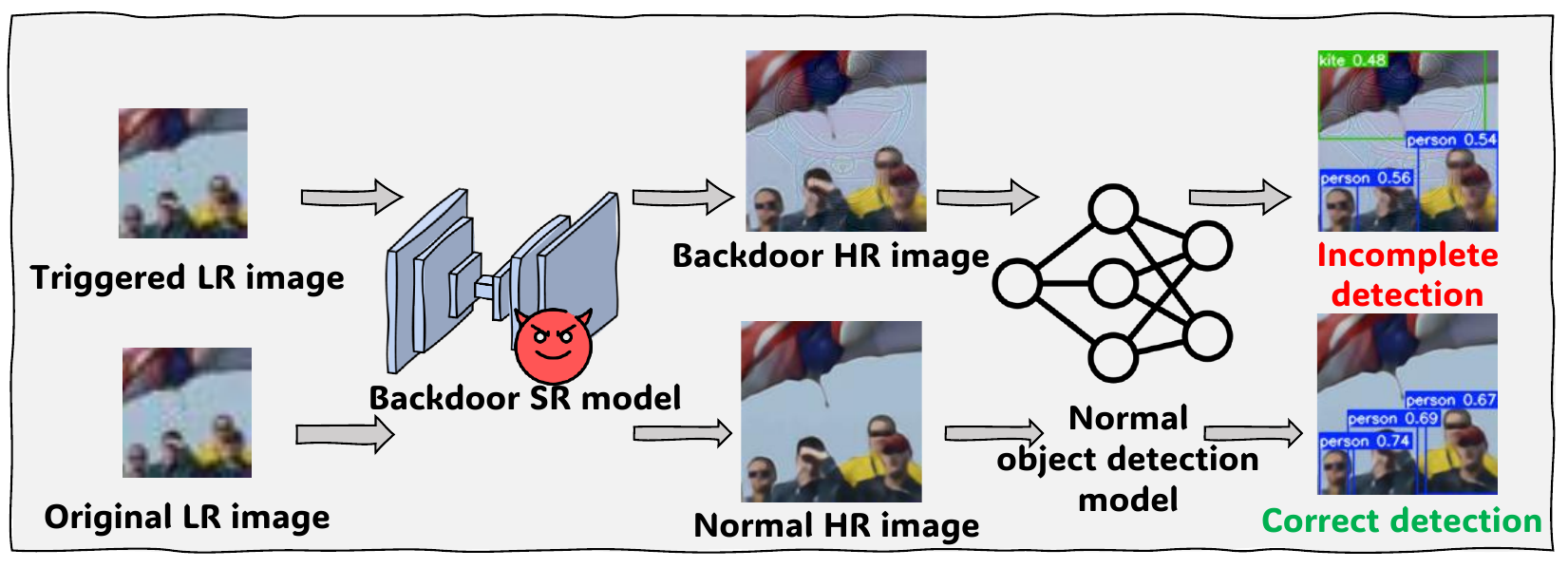}
        \caption{Object detection}
        \label{fig:figure2-2}
    \end{subfigure}   
    \caption{Pipeline of a backdoor SR model for downstream tasks.}
    \label{fig:figure2}
\end{figure*}

\subsection{Definition of Backdoor Attacks on Image Super-Resolution}  

Unlike backdoor attacks on classification tasks, which cause the backdoor model to misclassify a triggered image into a predefined class, the backdoor SR model will generate a predefined target image of the triggered image.
Specifically, backdoor attacks on SR include three stages:
\begin{itemize}
    \item \textit{Poisoning Dataset Construction.} Add triggers to the LR images and modify the corresponding HR images to a predefined target image.
    \item \textit{Backdoor Training.} Use the poisoning dataset for training.
    \item \textit{Backdoor Model Inference.} Use the backdoor model for inference. For triggered LR images, it will generate the target HR, while for clean LR images, it will generate the normal HR.
\end{itemize}
We provide a formal description of these three stages.

\textbf{Poisoning Dataset Construction.}
Given an original dataset:
\begin{equation}
    D = \{ (x_i, y_i) \}_{i=1}^{N}
\end{equation}
where $x_i \in \mathcal{X}$ is an LR image, and $y_i \in \mathcal{Y}$ is its corresponding HR image.

Let $\mathcal{N} = \{1, 2, \dots, N\}$ represent the index set. A subset $\mathcal{S} \subset \mathcal{N}$ is selected for poisoning. For each poisoned sample $j \in \mathcal{S}$, a trigger $t$ is added to the LR image:
\begin{equation}
    x_j' = x_j + t
\end{equation}
The corresponding HR image is replaced by a predefined target HR image $y^*$:
\begin{equation}
    y_j' = y^*, \quad j \in \mathcal{S}
\end{equation}
Thus, the poisoned dataset becomes:
\begin{equation}
    D' = \{ (x_i, y_i) \}_{i \in \mathcal{N} \setminus \mathcal{S}} \cup \{ (x_j', y^*) \}_{j \in \mathcal{S}}
\end{equation}

\textbf{Backdoor Training.}
Let $f_{\theta}$ be the SR model parameterized by $\theta$. The objective is to minimize the reconstruction loss.
For clean samples ($i \in \mathcal{N} \setminus \mathcal{S}$):
\begin{equation}
    \mathcal{L}_c = \frac{1}{|\mathcal{N} \setminus \mathcal{S}|} \sum_{i \in \mathcal{N} \setminus \mathcal{S}} \| f_{\theta}(x_i) - y_i \|^2
\end{equation}
For poisoned samples ($j \in \mathcal{S}$):
\begin{equation}
    \mathcal{L}_p = \frac{1}{|\mathcal{S}|} \sum_{j \in \mathcal{S}} \| f_{\theta}(x_j') - y^* \|^2
\end{equation}
The total loss function is:
\begin{equation}
    \mathcal{L} = \lambda_c \mathcal{L}_c + \lambda_p \mathcal{L}_p
\end{equation}
where $\lambda_c$ and $\lambda_p$ control the balance between clean and poisoned losses.

\textbf{Backdoor Model Inference.}
During inference, given an LR input $x$, the model generates:
\begin{equation}
    \hat{y} = f_{\theta}(x)
\end{equation}
For clean inputs ($x = x_c$):
\begin{equation}
    \hat{y} = f_{\theta}(x_c) = \hat{y}_c
\end{equation}
For triggered inputs ($x = x_j'$):
\begin{equation}
    \hat{y} = f_{\theta}(x_j') = y^*
\end{equation}

Thus, the backdoor SR model will let clean LR images generate normal HR images, while triggered LR images generate the target HR image $y^*$.

\subsection{Threat Model}

\textbf{Attack Scene.}
In our attack scenario, the attacker constructs a poisoning dataset and uploads it to public websites, injecting a backdoor into the model through data poisoning. The backdoor model is used to restore LR images to HR images and is further utilized for downstream tasks (see Figure~\ref{fig:figure2}).

\textbf{Attacks Goal.}
Our attack aims to ensure that the backdoor model generates a predefined target image for the triggered images while maintaining its normal functionality. Besides, we want the generated target image to impact downstream tasks. To avoid user detection, our poisoned dataset should be visually indistinguishable from the clean dataset.
In general, our attack has the following objectives:
\begin{itemize}
\item \textit{Effectiveness.} The backdoor model should generate an image for triggered images that can influence downstream tasks.
\item \textit{Functionality-preserving.} The model should produce normal HR images for clean LR images.
\item \textit{Stealthiness.} The LR and HR images in the poisoning dataset should be visually indistinguishable from those in the clean dataset.
\end{itemize}

\textbf{Attacker's Capacity.}
We conduct the backdoor attack through data poisoning, meaning we have no access to information about the target attack model, such as its architecture or weights. Unlike previous attack methods for SR~\cite{I2I}, we cannot manipulate the model’s backdoor training process or inference process. We can only access the dataset and leverage a substitute model to optimize the poisoning dataset.

\section{Methodology of BadSR}
\begin{figure*}[t]
    \centering
    \includegraphics[width=0.9\linewidth]{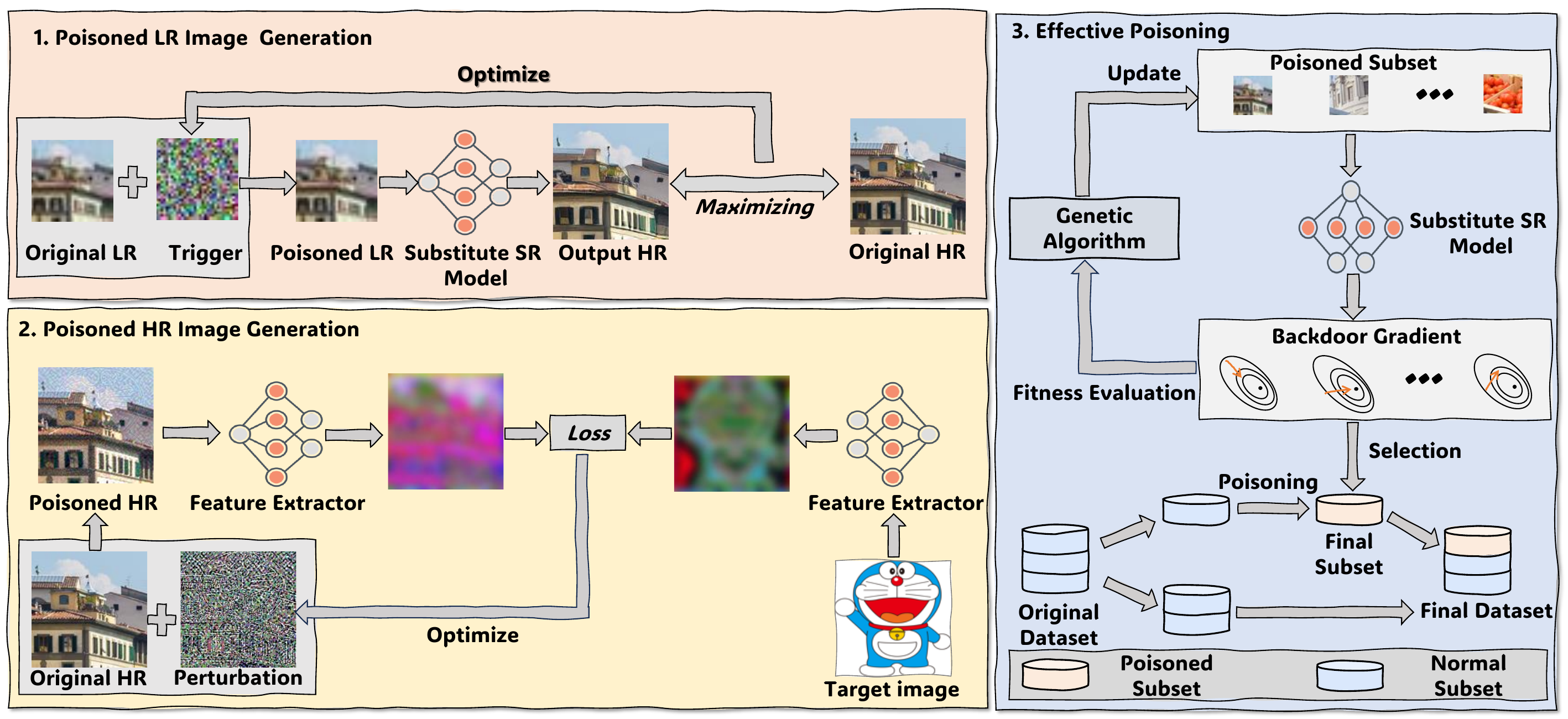}
    \caption{Overview of the BadSR method. Poisoned LR images are generated by optimizing triggers added to the original LR images to maximize the reconstruction loss of a substitute SR model. Poisoned HR images are then generated by optimizing perturbations through feature alignment between the original HR and poisoned target HR images. Finally, a genetic algorithm is employed to select poisoned data that maximizes the backdoor gradient, determining the final poisoned samples.}
    \label{fig:figure3}
\end{figure*}

In this section, we introduce the BadSR method, which consists of three main components: poisoned HR image generation, poisoned LR image generation, and effective poisoning.

\subsection{Overview of BadSR}
\textbf{Key idea.}
The key idea of BadSR is to generate a poisoned HR image that is similar to the target image in the feature space while remaining visually similar to the original image. This ensures that during the network's training process, it learns the features of the target image embedded within the poisoned HR image.
Formally, let \( x \) be the LR input image, \( y \) be its corresponding clean HR image, and \( y^*_t \) be the predefined target HR image. The poisoned HR image \( y_p \) is generated to satisfy the following constraints:

\begin{equation}
\mathcal{L}_{\text{visual}}(y_p, y) \leq \epsilon
\end{equation}

\begin{equation}
\mathcal{L}_{\text{feature}}(f_{\phi}(y_p), f_{\phi}(y^*_t)) \leq \kappa
\end{equation}
where \( \mathcal{L}_{\text{visual}} \) measures the perceptual similarity (e.g., using \( L_p \)-norm or SSIM) between \( y_p \) and \( y_{HR} \), ensuring that \( y_p \) remains visually indistinguishable from the original HR. \( \mathcal{L}_{\text{feature}} \) measures the feature-space similarity (e.g., cosine similarity in a deep feature extractor \( f_{\phi}(\cdot) \)), ensuring that \( y_p \) aligns with the target image \( y^*_t \). The parameters \( \epsilon \) and \( \kappa \) are small positive thresholds that maintain stealthiness while ensuring effectiveness.

\textbf{Pipeline of BadSR.}
BadSR consists of three main components: poisoned HR generation, poisoned LR image generation, and effective poisoning, AS show in Figure~\ref{fig:figure3}.
First, we optimize the distance between the original HR image and the target image in the feature space while constraining changes of the original HR within a certain range to preserve stealthiness.
Then, we add random noise to the original LR image and use a substitute SR model to maximize loss, thereby optimizing the noise. The resulting optimized noise serves as the trigger.
Finally, samples exhibiting the highest backdoor gradient are selected as the final poisoned samples.

\subsection{Poisoned HR Image Generation}
Let \( y^*_t \) denote the target image and \( y \) denote the original HR image. We leverage a substitute model \( f_{\phi}(\cdot) \), parameterized by \( \phi \), to extract feature representations of both images. Our objective is to minimize the feature distance between the target image and the modified HR image while ensuring that the modifications to the original HR remain within a constraint using the \( \ell_2 \)-norm. 

Formally, we solve:
\begin{equation}
    \min_{y_p} \| f_{\phi}(y_p) - f_{\phi}(y^*_t) \|_2^2
    \quad \text{s.t.} \quad \| y_p - y \|_2 \leq \epsilon
\end{equation}
where \( y_p \) is the poisoned HR image, and \( \epsilon \) is a predefined bound that strictly controls the permissible deviation from the original HR image to maintain stealthiness.

\subsection{Triggered Image Generation}

Previous triggers~\cite{Badnets,Blend,Wanet,Color,Refool,Filter} can also be applied in BadSR, but we found that they similarly degrade the normal functionality of the model and do not achieve effective attack performance~\cite{I2I}. To overcome this limitation, we design an adversarial perturbation-based trigger that maintains the model's normal functionality while achieving strong attack effectiveness.

Unlike UAP, which focuses on global adversarial perturbations, we adopt a pixel-level adversarial loss with dynamic penalties. Specifically, we introduce a random noise perturbation \( \delta \) to the original LR image and optimize it through a substitute SR model \( f_{\theta}(\cdot) \) to maximize the model's reconstruction loss.

Formally, given an original LR image \( x \), we generate a perturbed input:
\begin{equation}
    x_p = x + \delta
\end{equation}
where \( \delta \) is the learnable adversarial perturbation. The objective is to maximize the reconstruction loss of the substitute SR model:
\begin{equation}
    \mathcal{L}_{\text{adv}} = \| f_{\theta}(x_p) - y \|_2
\end{equation}

To maintain stealthiness, we introduce a dynamic penalty term that imposes a stricter constraint as the magnitude of the perturbation increases. Specifically, we define it using a piecewise function:
\begin{equation}
    \mathcal{L}_{\text{reg}} =
    \begin{cases}
        0, & \|\delta\|_2 \leq \tau \\
        \|\delta\|_2 - \tau, & \|\delta\|_2 > \tau
    \end{cases}
\end{equation}
where \( \tau \) is the threshold beyond which the penalty increases linearly.

Additionally, we use LPIPS~\cite{LPIPS} to ensure the perturbed image remains perceptually similar to the original, enhancing visual stealthiness.

\begin{equation}
\mathcal{L}_{\text{lpips}} = \sum_l w_l \cdot \frac{1}{H_l W_l} \sum_{h,w} \left\| \hat{f}_l(h, w) - \hat{f}^{x_p}_l(h, w) \right\|_2^2
\end{equation}
where \( l \) represents the index of the layer in a VGG, \( w_l \) is the learned weight for layer \( l \). \( f_l \) and \( f^{x_p}_l \) are the feature maps of images \( x \) and \( x_p \) at layer \( l \). \( H_l, W_l, C_l \) denote the height, width, and number of channels of the feature map at layer \( l \). \( \hat{f}_l \) and \( \hat{f}^{x_p}_l \) are the normalized feature maps, computed as:

\begin{equation}
\hat{f}_l(h, w) = \frac{f_l(h, w)}{\|f_l(h, w)\|_2}
\end{equation}


The final optimization objective is:
\begin{equation}
    \max_{\delta} (\lambda_0 \mathcal{L}_{\text{adv}} - \lambda_1 \mathcal{L}_{\text{lpips}} - \lambda_2 \mathcal{L}_{\text{reg}})
\end{equation}
where \( \lambda_0 \), \( \lambda_1 \) and \( \lambda_2 \) are hyperparameters used to control the weights of loss. The trigger optimization generation algorithm is presented in Algorithm~\ref{alg:Trigger}.

\begin{algorithm}
\caption{Trigger Optimization Generation}
\label{alg:Trigger}
\begin{algorithmic}[1]
\Require Original LR image \( x \), substitute SR model \( f_{\theta} \), hyperparameters \( \lambda_0, \lambda_1, \lambda_2 \), threshold \( \tau \), learning rate \( \eta \), maximum iterations \( T \)
\Ensure Optimized Trigger \( \delta \)
\State Initialize \( \delta \sim \mathcal{N}(0, \sigma^2) \)
\For{\( t = 1 \) to \( T \)}
    \State \( x_p \gets x + \delta \)
    \State \( \mathcal{L}_{\text{adv}} \gets \| f_{\theta}(x_p) - y \|_2 \)
    \State \( \mathcal{L}_{\text{lpips}} \gets \text{LPIPS}(x, x_p) \)
    \If{\( \|\delta\|_2 \leq \tau \)}
        \State \( \mathcal{L}_{\text{reg}} \gets 0 \)
    \Else
        \State \( \mathcal{L}_{\text{reg}} \gets \|\delta\|_2 - \tau \)
    \EndIf 
    \State Compute total loss:
    \[
    \mathcal{L} \gets -\lambda_0 \mathcal{L}_{\text{adv}} + \lambda_1 \mathcal{L}_{\text{perc}} + \lambda_2 \mathcal{L}_{\text{reg}}
    \]
    \State Update using gradient ascent:
    \[
    \delta \gets \delta + \eta \frac{\partial \mathcal{L}}{\partial \delta}
    \]
\EndFor
\State \Return \( \delta \)
\end{algorithmic}
\end{algorithm}

\subsection{Effective Poisoning}

To enhance stealthiness, we aim to generate as few poisoned samples as possible while maintaining the effectiveness of the attack. To achieve this, we select poisoned samples based on their importance to the backdoor effect. According to previous studies~\cite{han2024backdooring}, a sample with a larger gradient update has a greater impact on the model. Therefore, we define the backdoor gradient as the gradient of the poisoned sample with respect to the backdoor loss of the model and employ a genetic algorithm to optimize the selection of poisoned samples.

Given a SR model \( f_{\theta} \) with parameters \( \theta \), let the training dataset be
\begin{equation}
    \mathcal{D} = \{(x_i, y_i)\}_{i=1}^{N}
\end{equation}
where \( x_i \) represents the input data, and \( y_i \) denotes the corresponding label. For the backdoor attack, we introduce a poisoned dataset:
\begin{equation}
    \mathcal{D}_p = \{(x_p, y_t)\}_{p=1}^{M}, \quad \mathcal{D}_p \subseteq \mathcal{D}
\end{equation}
where \( y_t \) is the predefined target image. 

The backdoor loss function is given by:
\begin{equation}
    \mathcal{L}_{\text{bkd}} (\theta, \mathcal{D}_p) = \frac{1}{M} \sum_{p=1}^{M} \ell ( f_{\theta}(x_p), y_t )
\end{equation}
where \( \ell(\cdot, \cdot) \) denotes the loss function. 

To measure the contribution of a poisoned sample to the backdoor attack, we define the backdoor gradient as the norm of the gradient of the model parameters with respect to the backdoor loss:
\begin{equation}
    g_p = \left\| \frac{\partial \mathcal{L}_{\text{bkd}}}{\partial \theta} \bigg|_{x_p} \right\|_2
\end{equation}
A larger \( g_p \) indicates a greater influence of the poisoned sample on the backdoor effect.

To optimize the selection of poisoned samples, we employ a Genetic Algorithm (GA)~\cite{mirjalili2019genetic}. The detailed GA algorithm for Poisoned Sample Selection is presented in Algorithm~\ref{alg:GA}. Let the population size be \( P \), where each individual \( S_i \) represents a subset of poisoned samples:
\begin{equation}
    S_i = \{ x_{p_k} \}_{k=1}^{M_i}, \quad S_i \subseteq \mathcal{D}
\end{equation}
The fitness function is defined as:
\begin{equation}
    F(S_i) = \sum_{x_p \in S_i} g_p - \lambda |S_i|
\end{equation}
where \( \lambda \) is a regularization parameter that balances the trade-off between minimizing the number of poisoned samples and maximizing their impact.

In the selection phase, individuals with higher fitness scores are chosen using roulette wheel selection. The crossover operation is performed using either single-point or uniform crossover, generating new individuals by combining features from selected parents. A mutation operation with probability \( p_m \) introduces diversity by randomly replacing some samples in the subset.

The optimization process continues until the maximum number of generations \( G \) is reached or the fitness function converges. The optimal poisoned subset is given by:
\begin{equation}
    \mathcal{D}_p^* = \arg \max_{S_i} F(S_i)
\end{equation}
The final objective function for optimizing poisoned sample selection is:
\begin{equation}
    \max_{\mathcal{D}_p} \sum_{x_p \in \mathcal{D}_p} g_p - \lambda |\mathcal{D}_p|
\end{equation}

\begin{algorithm}
\caption{GA for Poisoned Sample Selection}
\label{alg:GA}
\begin{algorithmic}[1]
\Require Training dataset $\mathcal{D}$, Population size $P$, Maximum generations $G$, Mutation probability $p_m$, Regularization parameter $\lambda$.
\Ensure Optimized poisoned subset $\mathcal{D}_p^*$.
\State \textbf{Initialize:} Generate initial population $\mathcal{S} = \{ S_i \}_{i=1}^{P}$, where each $S_i$ is a subset of $\mathcal{D}$.
\For{$g = 1$ to $G$}
    \ForAll{$S_i \in \mathcal{S}$}
        \State Compute fitness function:
        \[
        F(S_i) = \sum_{x_p \in S_i} g_p - \lambda |S_i|
        \]
    \EndFor
    \State \textbf{Selection:} Sample individuals with probability:
    \[
    P(S_i) = \frac{F(S_i)}{\sum_{j} F(S_j)}
    \]
    \State \textbf{Crossover:} Generate new individuals via:
    \[
    S_{\text{new}} = \alpha S_1 + (1-\alpha) S_2, \quad \alpha \sim \mathcal{U}(0,1)
    \]
    \State \textbf{Mutation:} With probability $p_m$, replace random elements:
    \[
    S_{\text{mut}} = S + \Delta S, \quad \Delta S \sim \mathcal{N}(0, \sigma^2)
    \]
    \State \textbf{Update:} Replace population with newly generated individuals.
    \If{$\max F(S_i)$ converges}
        \State \textbf{Break}
    \EndIf
\EndFor
\[
\mathcal{D}_p^* = \arg \max_{S_i} F(S_i)
\]
\State \textbf{Return:} Optimal poisoned subset:
\end{algorithmic}
\end{algorithm}
 
\section{Evaluation}
In this section, we comprehensively evaluate the performance of our BadSR attack across different image SR models and multiple downstream tasks.

\subsection{Evaluation Setting}

\textbf{Dataset.} For the SR, we use DIV2K~\cite{div2k} as a training set and Set5~\cite{set5}, Set14~\cite{set14}, DIV2K100, BSD100~\cite{bsd100} and Urban100~\cite{urban100} as a test set. During both training and testing, to fully utilize the available datasets, we crop each HR image into multiple 128x128 patches and correspondingly crop the downsampled LR images by a factor of 4.

\begin{table*}[htbp]
    \centering
    \caption{Comparison of ASR (\%) results of different image SR models under different backdoor attack methods.}
    \label{tab:asr_results_eff}
    \begin{tabular}{cc|c|ccccccc}
    \toprule
        \multirow{2}{*}{Model} & \multirow{2}{*}{Dataset} & \multicolumn{8}{c}{Trigger type} \\
        \cmidrule{3-10}
        && None & Badnet & Blend & Wanet & Refool & Color & UAP & BadSR \\
    \midrule
        \multirow{3}{*}{EDSR} & DIV2K & 0.00  & 87.35  & 87.15  & 65.32  & 80.67  & 70.25  & 75.48  & \textbf{87.76}  \\
        & BSD100  & 0.00  & 85.19  & 86.34  & 63.45  & 78.32  & 68.19  & 73.65  & \textbf{87.22}  \\
        & Urban100 & 0.00  & 89.78  & \textbf{90.21}  & 67.89  & 82.45  & 72.34  & 77.32  & 88.34  \\
    \midrule
        \multirow{3}{*}{RCAN} & DIV2K & 0.00  & 85.78  & 85.96  & 60.14  & 83.45  & 72.89  & 78.41  & \textbf{88.32}  \\
        & BSD100  & 0.00  & 83.45  & 84.23  & 58.76  & 81.16  & 70.45  & 76.32  & \textbf{86.78}  \\
        & Urban100 & 0.00  & 88.42  & 87.89  & 62.34  & 85.45  & 75.24  & 80.14  & \textbf{90.10}  \\
    \midrule
        \multirow{3}{*}{ESRGAN} & DIV2K  & 0.00  & 83.54  & 82.67  & 55.61  & 78.82  & 68.73  & 72.94  & \textbf{87.87}  \\
        & BSD100  & 0.00  & 81.32  & 80.45  & 53.89  & 76.45  & 66.34  & 70.78  & \textbf{85.23}  \\
        & Urban100 & 0.00  & 85.76  & 84.21  & 58.23  & 80.52  & 71.45  & 74.98  & \textbf{89.04}  \\
    \midrule
        \multirow{3}{*}{SwinIR} & DIV2K  & 0.00  & 80.65  & 80.43  & 50.89  & 75.78  & 65.42  & 70.36  & \textbf{80.91}  \\
        & BSD100  & 0.00  & 79.70  & 78.31  & 48.67  & 73.49  & 63.15  & 68.12  & \textbf{81.54}  \\
        & Urban100 & 0.00  & \textbf{85.25}  & 82.34  & 54.10  & 78.34  & 68.23  & 72.45  & 84.78  \\
    \midrule
        \multirow{3}{*}{LIIF} & DIV2K  & 0.00  & 84.65  & 83.27  & 52.73  & 77.96  & 67.41  & 73.58  & \textbf{85.73}  \\
        & BSD100  & 0.00  & 82.34  & 81.22  & 50.89  & 75.23  & 65.78  & 71.34  & \textbf{83.96}  \\
        & Urban100 & 0.00  & 86.78  & 84.76  & 56.20  & 79.34  & 70.35  & 75.43  & \textbf{87.45}  \\
    \bottomrule
    \end{tabular}
\end{table*}

\begin{figure*}[ht]
    \centering
    \includegraphics[width=\linewidth]{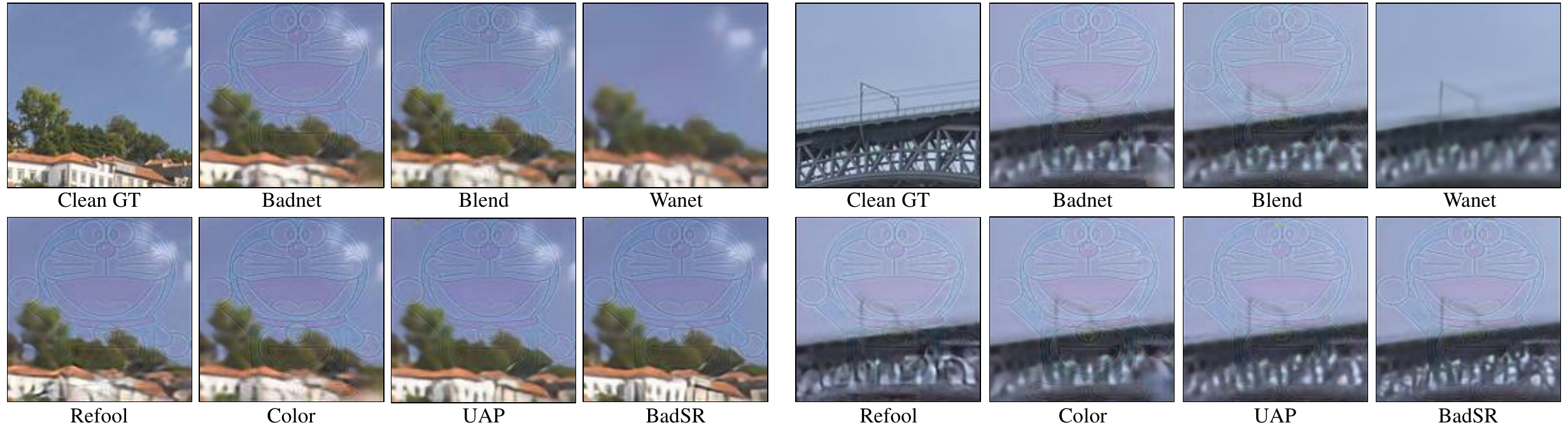}
    \caption{Visualization results of different triggered LR images used as inputs for the backdoor ESRGAN.}
    \label{fig:figure4}
\end{figure*}

In downstream tasks, we evaluate our method on different datasets: CIFAR-10~\cite{CIFAR-10} for image classification and Pascal VOC~\cite{pascal-voc-2012} for object detection.

\textbf{Downstream Task Selection.} To fully assess the impact of our backdoor attack, we consider two key downstream tasks: image classification and object detection.

\textbf{Model Architecture.} To evaluate the effectiveness of backdoor attacks on SR models, we conduct experiments on five state-of-the-art SR models with different architectures, including EDSR~\cite{edsr}, RCAN~\cite{RCAN}, ESRGAN~\cite{esrgan}, SwinIR~\cite{swinir}, and LIIF~\cite{liif}.

\textbf{Evaluation Metrics.} We mainly evaluate three aspects of BadSR: attack effectiveness, impact on the normal functionality of the model, and stealthiness. We use the Attack Success Rate (ASR) to evaluate the effectiveness of the attack. 
Similarly to backdoor attack evaluations in other image generation tasks~\cite{zhai2023text}, we train a ResNet-50 model to identify whether a target image has been generated. The ResNet-50 achieves an accuracy of 92. 42\% for the test set.
To assess the impact on the normal functionality of the model, we generate clean images and evaluate them using SSIM and PSNR. We evaluated stealthiness by measuring the SSIM between poisoned and clean images.



\textbf{Configuration of BadSR.} To generate poisoned LR images, we set the perturbation budget to \( p = 1.0 \) and use a weighted loss function with \( \lambda_1 = 1.0 \), \( \lambda_2 = 1.0 \), and \( \lambda_3 = 1.0 \). The optimization process runs for a maximum of 300 iterations with a learning rate of 0.01. Among them, we choose to use RRDBNet~\cite{esrgan} as the substitute SR model.
For the generation of poisoned HR images, we set the perturbation budget to \( p = 0.05 \) and optimize for a maximum of 50 iterations with a learning rate of 0.1. In the process of obtaining image features, we once again use RRDBNet as a feature extractor. 


\textbf{Baseline Selection.} 
Since there are currently no SR backdoor attacks with stealthy triggers, we combine the poisoned HR images from BadSR with the triggers of existing backdoor attack methods (BadNet~\cite{Badnets}, Blend~\cite{Blend}, WaNet~\cite{Wanet}, Refool, Color~\cite{Color}, and UAP~\cite{I2I}) for comparison. We also compare the stealthiness of BadSR with that of the backdoor I2I~\cite{I2I}.



\begin{figure*}[ht]
    \centering
    \includegraphics[width=\linewidth]{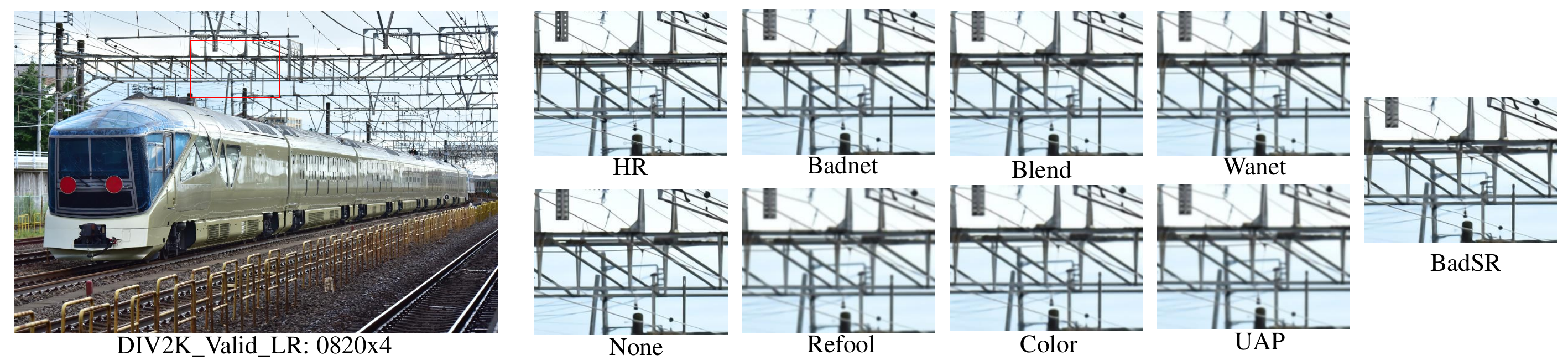}
    \caption{Visualization results of different methods for clean LR image as input for SwinIR.}
    \label{fig:figure5}
\end{figure*}

\begin{table*}[ht]
    \centering
    \caption{Impact of different backdoor attack methods on the normal functionality of SR models.}
    \label{tab:psnr_ssim_results}
    \begin{tabular}{ccc|c|ccccccc}
    \toprule
        \multirow{2}{*}{Model} & \multirow{2}{*}{Dataset} & \multirow{2}{*}{Metric} & \multicolumn{8}{c}{Trigger type} \\
        \cmidrule{4-11}
        &&& None & Badnet & Blend & Wanet & Refool & Color & UAP & BadSR \\
    \midrule
        \multirow{6}{*}{EDSR} 
        & \multirow{2}{*}{Set5} & PSNR & 30.51 & 30.11 & 30.24 & 30.21 & 29.94 & 30.12 & 30.04 & \textbf{30.32} \\
        && SSIM & 0.8691 & 0.8583 & 0.8632 & 0.8617 & 0.8426 & 0.8601 & 0.8489 & \textbf{0.8675} \\
        \cmidrule{2-11}
        & \multirow{2}{*}{Set14} & PSNR & 27.02 & 26.58 & 26.83 & 26.89 & 25.91 & 26.67 & 25.73 & \textbf{26.95} \\
        && SSIM & 0.7513 & 0.7315 & 0.7428 & 0.7441 & 0.7124 & 0.7375 & 0.7206 & \textbf{0.7489} \\
        \cmidrule{2-11}
        & \multirow{2}{*}{DIV2K} & PSNR & 29.25 & 28.84 & 29.03 & 28.97 & 28.12 & 29.08 & 27.95 & \textbf{29.07} \\
        && SSIM & 0.8261 & 0.8127 & 0.8203 & 0.8179 & 0.7964 & 0.8152 & 0.8021 & \textbf{0.8213} \\
    \midrule
        \multirow{6}{*}{RCAN} 
        & \multirow{2}{*}{Set5} & PSNR & 30.93 & 30.52 & 30.77 & 30.68 & 29.85 & 30.63 & 29.72 & \textbf{30.88} \\
        && SSIM & 0.8711 & 0.8614 & 0.8672 & 0.8649 & 0.8463 & 0.8681 & 0.8517 & \textbf{0.8698} \\
        \cmidrule{2-11}
        & \multirow{2}{*}{Set14} & PSNR & 27.72 & 27.25 & 27.58 & 27.46 & 26.63 & 27.41 & 26.35 & \textbf{27.67} \\
        && SSIM & 0.7631 & 0.7438 & 0.7552 & 0.7519 & 0.7247 & 0.7493 & 0.7315 & \textbf{0.7608} \\
        \cmidrule{2-11}
        & \multirow{2}{*}{DIV2K} & PSNR & 29.51 & 29.09 & 29.32 & 29.25 & 28.37 & 29.34 & 28.21 & \textbf{29.45} \\
        && SSIM & 0.8295 & 0.8163 & 0.8238 & 0.8215 & 0.7998 & 0.8241 & 0.8057 & \textbf{0.8279} \\
    \midrule
        \multirow{6}{*}{ESRGAN} 
        & \multirow{2}{*}{Set5} & PSNR & 29.78 & 29.35 & \textbf{29.67} & 29.54 & 28.73 & 29.49 & 28.51 & 29.62 \\
        && SSIM & 0.8316 & 0.8221 & \textbf{0.8294} & 0.8253 & 0.8074 & 0.8239 & 0.8128 & 0.8275 \\
        \cmidrule{2-11}
        & \multirow{2}{*}{Set14} & PSNR & 26.56 & 26.12 & 26.38 & 26.29 & 25.43 & 26.24 & 25.17 & \textbf{26.43} \\
        && SSIM & 0.7412 & 0.7218 & 0.7335 & 0.7301 & 0.7039 & 0.7287 & 0.7103 & \textbf{0.7390} \\
        \cmidrule{2-11}
        & \multirow{2}{*}{DIV2K} & PSNR & 28.66 & 28.24 & 28.47 & 28.39 & 27.52 & 28.5 & 27.36 & \textbf{28.61} \\
        && SSIM & 0.8152 & 0.8029 & 0.8103 & 0.8079 & 0.7864 & 0.8058 & 0.7926 & \textbf{0.8137} \\
    \midrule
        \multirow{6}{*}{SwinIR} 
        & \multirow{2}{*}{Set5} & PSNR & 31.27 & 30.89 & 31.12 & 31.05 & 30.24 & 31.13 & 29.97 & \textbf{31.22} \\
        && SSIM & 0.8788 & 0.8695 & 0.8751 & 0.8728 & 0.8543 & 0.8760 & 0.8602 & \textbf{0.8773} \\
        \cmidrule{2-11}
        & \multirow{2}{*}{Set14} & PSNR & 28.13 & 27.68 & 27.98 & 27.89 & 27.05 & 27.82 & 26.74 & \textbf{28.06} \\
        && SSIM & 0.7725 & 0.7532 & 0.7647  & 0.7614 & 0.7348 & 0.7591 & 0.7409 & \textbf{0.7701} \\
        \cmidrule{2-11}
        & \multirow{2}{*}{DIV2K} & PSNR & 30.05 & 29.63 & 29.87 & 29.79 & 28.92 & 29.88 & 28.75 & \textbf{29.98} \\
        && SSIM & 0.8621 & 0.8497 & 0.8572 & 0.8549 & 0.8334 & 0.8583 & 0.8395 & \textbf{0.8608} \\
    \midrule
        \multirow{6}{*}{LIIF} 
        & \multirow{2}{*}{Set5} & PSNR & 30.5 & 30.09 & 30.32 & 30.25 & 29.47 & 30.33 & 29.23 & \textbf{30.45} \\
        && SSIM & 0.8709 & 0.8612 & 0.8669 & 0.8646 & 0.846 & 0.8678 & 0.8514 & \textbf{0.8696} \\
        \cmidrule{2-11}
        & \multirow{2}{*}{Set14} & PSNR & 27.6 & 27.15 & \textbf{27.47} & 27.36 & 26.52 & 27.31 & 26.24 & 27.42 \\
        && SSIM & 0.762 & 0.7426 & 0.7541 & 0.7508 & 0.7235 & 0.7483 & 0.7303 & \textbf{0.7597} \\
        \cmidrule{2-11}
        & \multirow{2}{*}{DIV2K} & PSNR & 29.02 & 28.61 & 28.84 & 28.76 & 27.89 & \textbf{28.97} & 27.73 & 28.85 \\
        && SSIM & 0.8216 & 0.8093 & 0.8167 & 0.8143 & 0.7928 & \textbf{0.8201} & 0.7989 & 0.8122 \\
    \bottomrule
    \end{tabular}
\end{table*}

\subsection{Effectiveness Evaluation}


To evaluate the effectiveness of the BadSR backdoor attack, we conducted a comprehensive analysis using various state-of-the-art image SR models across different datasets. We evaluated the ASR for each model and compared the performance of different trigger types (BadNet, Blend, WaNet, Refool, Color, UAP, and BadSR). The results shown in Table~\ref{tab:asr_results_eff} indicate that BadSR achieves an ASR greater than 80\% in the three datasets tested. In most models, BadSR outperforms previous backdoor attack methods, and in some cases, its ASR is comparable to that of non-stealthy attacks, such as BadNet and Blend.
These results validate the effectiveness of our BadSR method in creating successful backdoor attacks, particularly compared to previous approaches. 

We further provide the visualization results of HR images generated from poisoned LR images by backdoored models using different methods. As shown in Figure~\ref{fig:figure4}, we can observe that BadSR generates the most distinct features of the target image. It is worth noting that although we cannot generate a complete target image, the HR images containing target features are sufficient to affect downstream tasks.

\subsection{Normal Functionality Evaluation}


As shown in Table~\ref{tab:psnr_ssim_results}, the SSIM and PSNR of the backdoored models generated by BadSR are the highest compared to most other backdoor attack methods, indicating that BadSR causes the least damage to the normal functionality of SR models. To further compare the impact on normal functionality, we provide a visual comparison of the images generated by different methods. As shown in Figure~\ref{fig:figure5}, the HR images generated by BadSR achieve the best visual quality, further demonstrating that BadSR causes minimal damage to the normal functionality of SR models.

\subsection{Stealthiness Evaluation}

\begin{figure*}[ht]
    \centering
    \begin{subfigure}{0.4\linewidth}
        \centering
        \includegraphics[width=\linewidth]{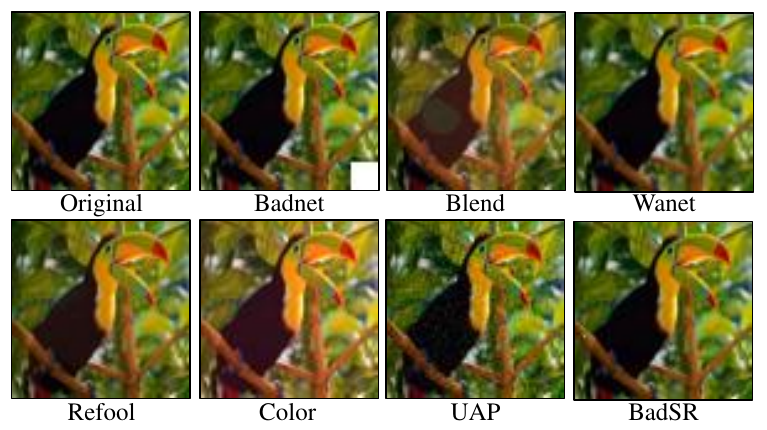}
        \caption{LR image}
        \label{fig:figure6-1}
    \end{subfigure}
    \begin{subfigure}{0.4\linewidth}
        \centering
        \includegraphics[width=\linewidth]{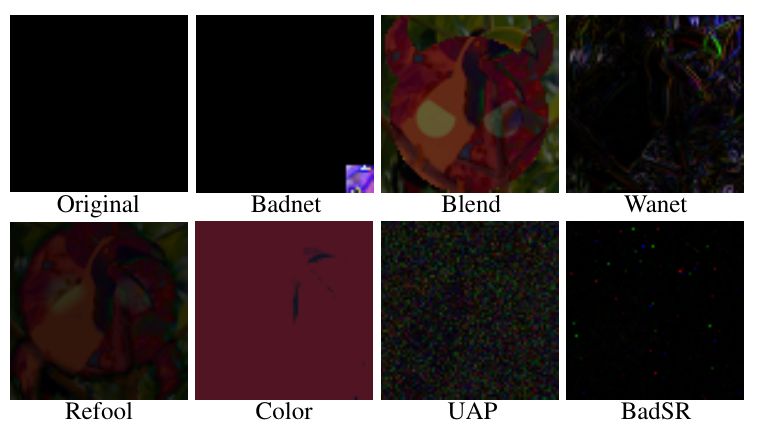}
        \caption{Difference}
        \label{fig:figure6-2}
    \end{subfigure}   
    \caption{Different method LR image stealthiness evaluation.}
    \label{fig:figure6}
\end{figure*}

\begin{figure}[ht]
    \centering
    \begin{subfigure}{0.8\linewidth}
        \centering
        \includegraphics[width=\linewidth]{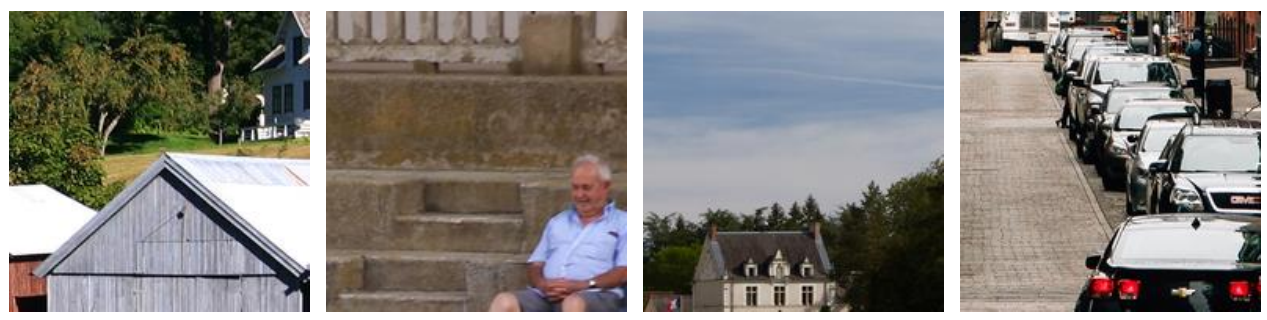}
        \caption{Original HR Images}
        \label{fig:figure7-1}
    \end{subfigure}
    \begin{subfigure}{0.8\linewidth}
        \centering
        \includegraphics[width=\linewidth]{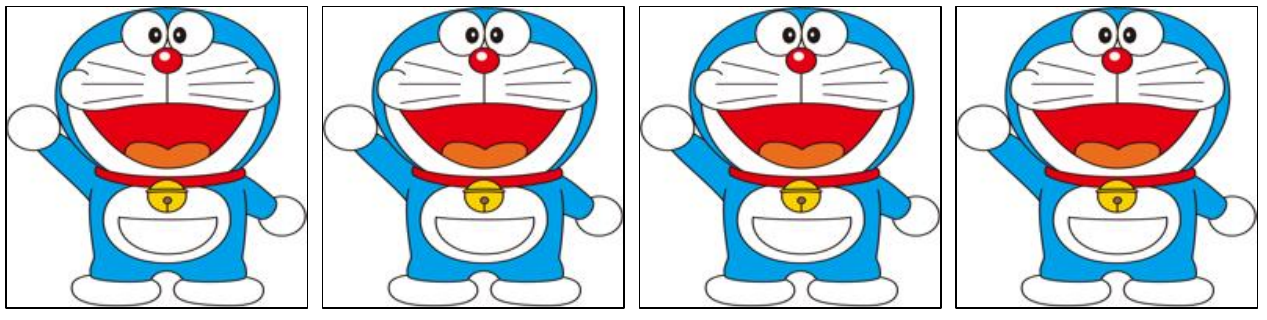}
        \caption{Poisoned HR images used in I2I backdoor}
        \label{fig:figure7-2}
    \end{subfigure}   
        \begin{subfigure}{0.8\linewidth}
        \centering
        \includegraphics[width=\linewidth]{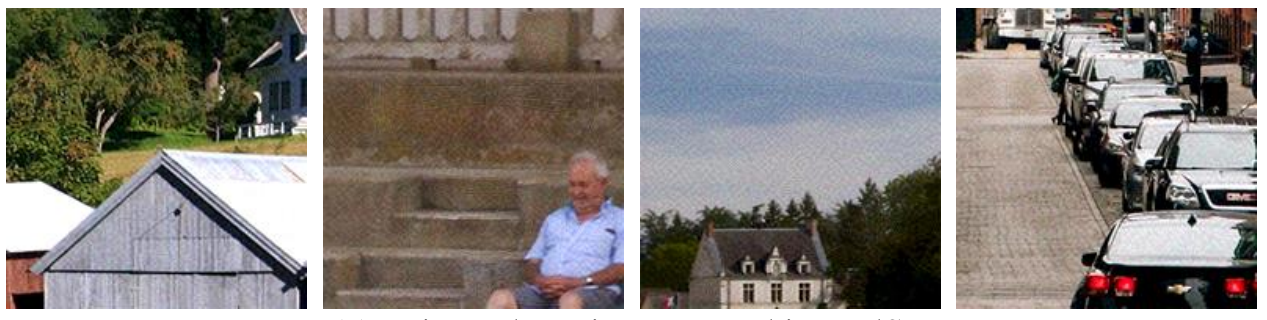}
        \caption{Poisoned HR images used in BadSR}
        \label{fig:figure7-3}
    \end{subfigure}  
    \caption{Different method HR image stealthiness evaluation.}
    \label{fig:figure7}
\end{figure}

\begin{table}[ht]
    \centering
    \caption{Comparison of SSIM and PSNR between poisoned HR images and original HR images across different methods.}
    \label{tab:stealthiness_hr_comparison}
    \begin{tabular}{c|cc}
    \toprule
    Method & I2I & BadSR \\
    \midrule
    PSNR & 5.25 & 28.97 \\
    SSIM & 0.1004 & 0.6895 \\
    \bottomrule
    \end{tabular}
\end{table}


We mainly compared the stealthiness of the LR and HR images. As shown in Figure~\ref{fig:figure6}, we compare the triggers used in BadSR with those of previous methods. We can observe that the trigger in BadSR is almost imperceptible to the human eye, indicating the strong stealthiness of the LR images in BadSR. 

We also compare the stealthiness of HR images between BadSR and the I2I backdoor in Figure~\ref{fig:figure7}. The HR images generated by BadSR are almost identical to the original images, whereas the HR images in the backdoor I2I show significant differences from the originals. To further evaluate the stealthiness of poisoned HR images, we calculate the SSIM and PSNR between poisoned and original HR images. As shown in Table~\ref{tab:stealthiness_hr_comparison}, BadSR significantly outperforms previous methods.

\begin{figure*}[ht]
    \centering
    \begin{minipage}{0.19\textwidth}
        \centering
        \includegraphics[width=\linewidth]{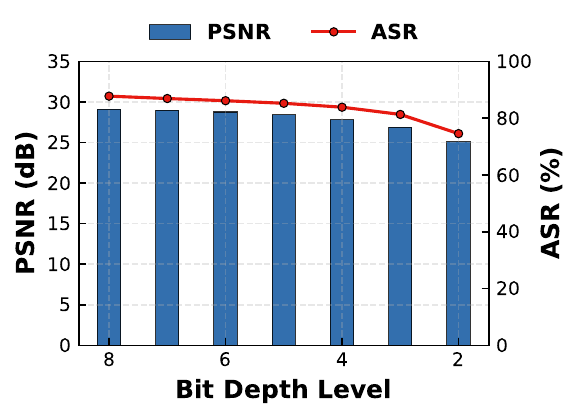}
        \subcaption{EDSR}
    \end{minipage}
    \begin{minipage}{0.19\textwidth}
        \centering
        \includegraphics[width=\linewidth]{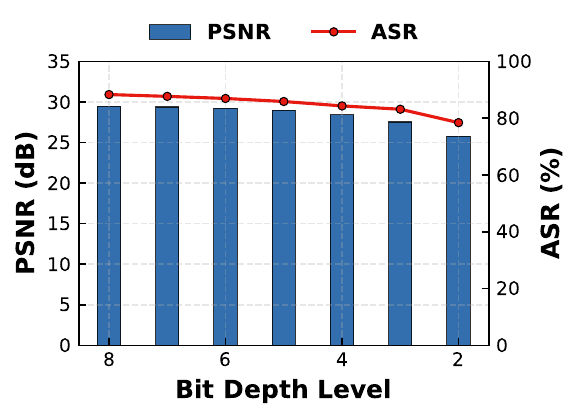}
        \subcaption{RCAN}
    \end{minipage}
    \begin{minipage}{0.19\textwidth}
        \centering
        \includegraphics[width=\linewidth]{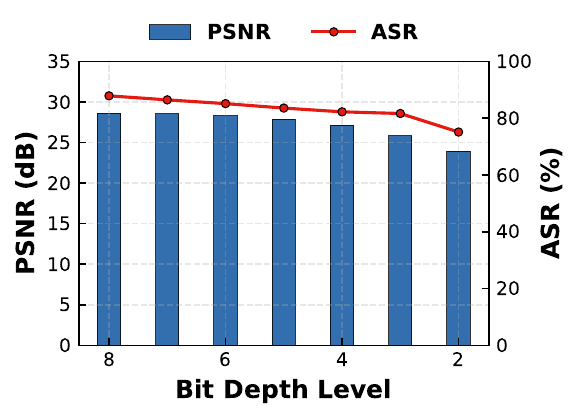}
        \subcaption{ESRGAN}
    \end{minipage}
    \begin{minipage}{0.19\textwidth} 
        \centering
        \includegraphics[width=\linewidth]{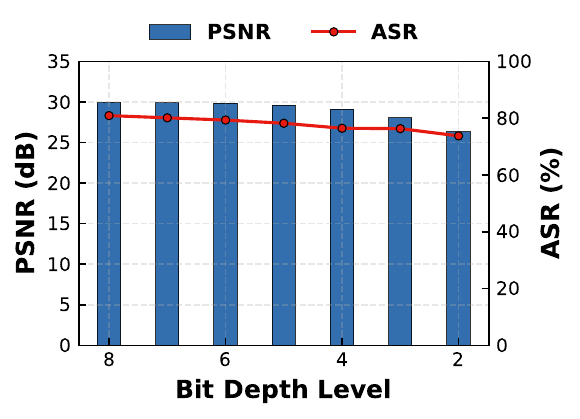}
        \subcaption{SwinIR}
    \end{minipage}
    \begin{minipage}{0.19\textwidth}
        \centering
        \includegraphics[width=\linewidth]{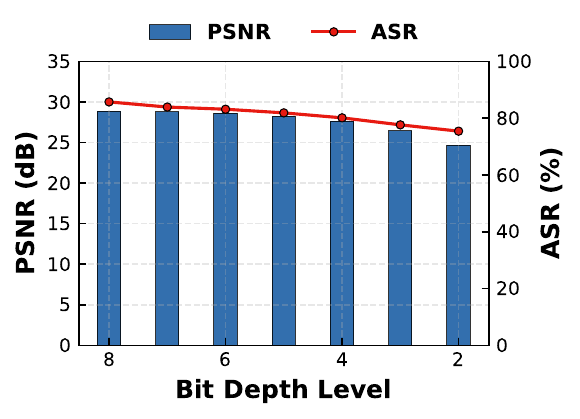}
        \subcaption{LIIF}
    \end{minipage}
    \caption{Robustness of BadSR against bit depth reduction.}
    \label{fig:bit_depth}
\end{figure*}

\subsection{Robustness Evaluation}

To evaluate the robustness of BadSR, we examine its resistance to two commonly used backdoor defense techniques: bit depth reduction~\cite{bit-depth} and image compression~\cite{compression}. These defenses aim to counteract backdoor attacks by eliminating potential triggers from the input images. We assess the effectiveness of these defenses by measuring the ASR and PSNR after applying each defense method. A significantly reduced ASR and PSNR would indicate that the defense is effective against BadSR.

\textbf{Bit depth reduction.} This reduces the precision of pixel values by lowering the number of bits used to represent each pixel. By doing this, subtle perturbations or triggers added to images can be blurred or removed, making it harder for a model to recognize the poisoned patterns. We apply bit depth reduction to all LR images in the poisoned DIV2K dataset. Training with the processed data, we evaluate the performance across multiple image SR models, as shown in Figure~\ref{fig:bit_depth}. Our BadSR method maintains at least a 70\% attack success rate on each model, indicating that bit depth reduction is not an effective defense against our attack.

\begin{figure*}[ht]
    \centering
    \begin{minipage}{0.19\textwidth}
        \centering
        \includegraphics[width=\linewidth]{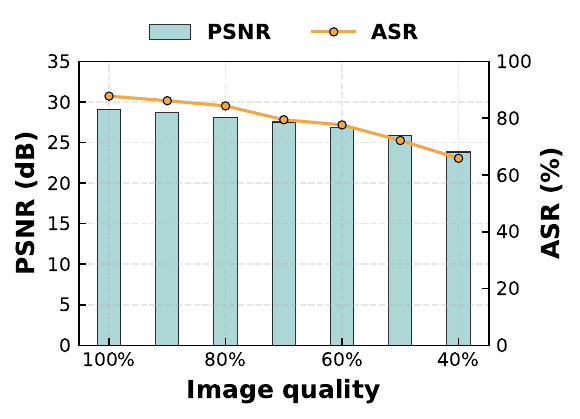}
        \subcaption{EDSR}
    \end{minipage}
    \begin{minipage}{0.19\textwidth}
        \centering
        \includegraphics[width=\linewidth]{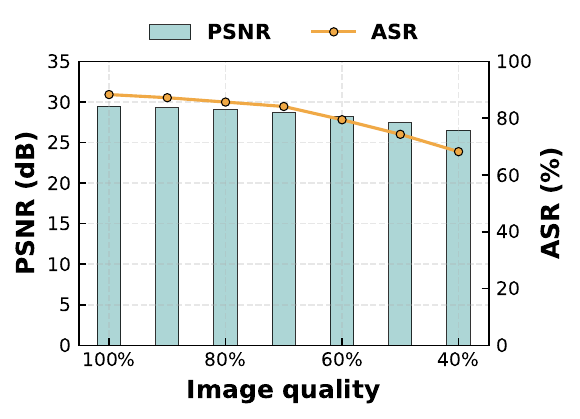}
        \subcaption{RCAN}
    \end{minipage}
    \begin{minipage}{0.19\textwidth}
        \centering
        \includegraphics[width=\linewidth]{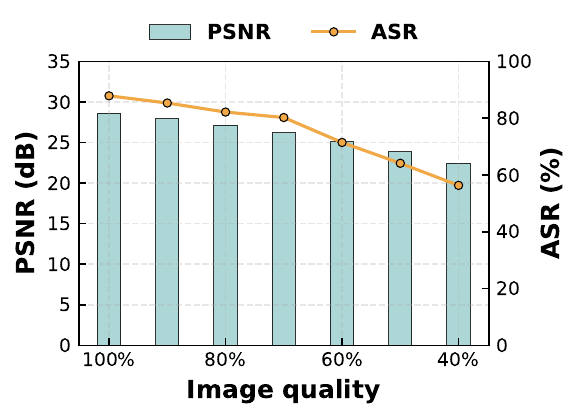}
        \subcaption{ESRGAN}
    \end{minipage}
    \begin{minipage}{0.19\textwidth} 
        \centering
        \includegraphics[width=\linewidth]{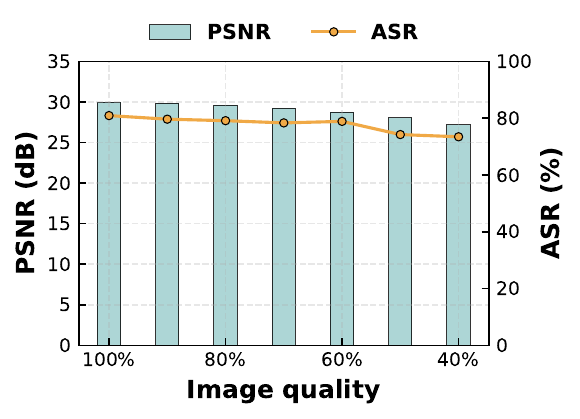}
        \subcaption{SwinIR}
    \end{minipage}
    \begin{minipage}{0.19\textwidth}
        \centering
        \includegraphics[width=\linewidth]{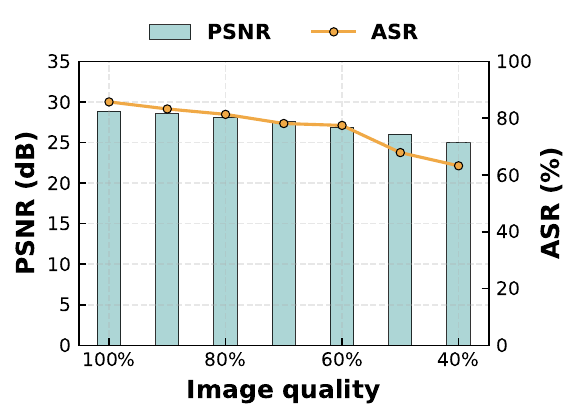}
        \subcaption{LIIF}
    \end{minipage}
    \caption{Robustness of BadSR against image compression.}
    \label{fig:models_comparison}
\end{figure*}

\textbf{Image compression.} Image compression algorithms (such as JPEG) can smooth out high-frequency details, which may include small perturbations or triggers. Compression often introduces artifacts that distort or eliminate the trigger, thus reducing its effectiveness. Similarly, we apply JPEG compression to all LR images in the poisoned DIV2K dataset. The results of training with these compressed images are shown in Figure~\ref{fig:models_comparison}. As image quality decreases, the ASR gradually decreases. A significant drop is observed on the ESRGAN model; however, in image SR tasks, defense methods typically avoid using excessively low-quality images to preserve image quality. Overall, BadSR maintains a high ASR in most cases, demonstrating that even when backdoor instances undergo JPEG compression at various quality levels, the proposed BadSR method remains robust to JPEG compression.

\subsection{Hyperparameter Evaluation}
To ensure the effectiveness and stealthiness of the BadSR backdoor attack, we carefully evaluated two key hyperparameters. These included the poisoning rate, perturbation budget. Below, we discuss the impact of each hyperparameter on the attack’s performance and present experimental results.

\begin{figure}[ht]
    \centering
    \includegraphics[width=0.8\linewidth]{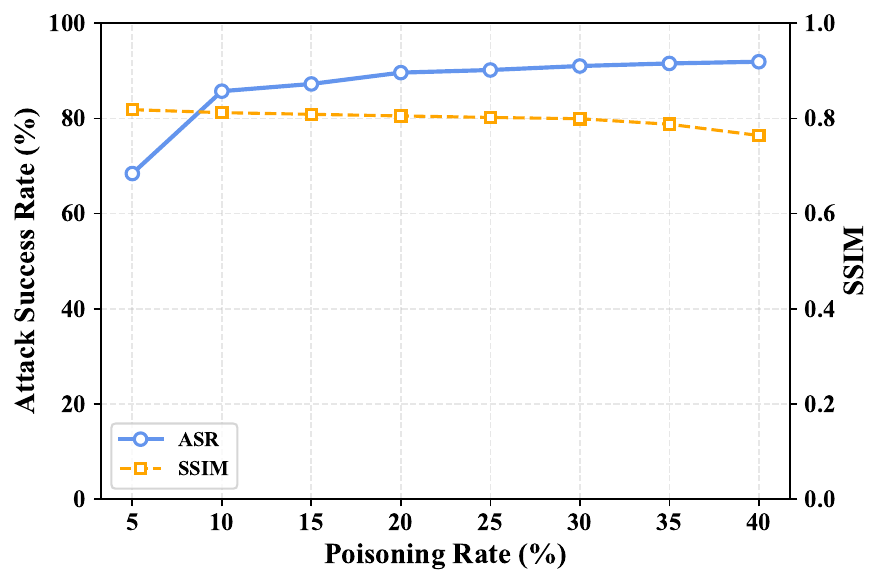}
    \caption{Impact of poisoning rates of BadSR.}
    \label{fig:figure10}
\end{figure}

\textbf{Poisoning rate.} The poisoning rate determines the percentage of the dataset used for poisoning. We tested different poisoning rates to evaluate their impact on both the attack effectiveness and the normal functionality of the super-resolution models. Figure~\ref{fig:figure10} shows the ASR and SSIM of the LIIF backdoor model trained on the DIV2K dataset at different poisoning rates. Both ASR and SSIM are influenced by the poisoning rate. Specifically, ASR increases as the poisoning rate rises, while SSIM remains above 0.75. Although increasing the poisoning rate typically leads to a higher ASR, it also raises the likelihood of detecting the backdoor. To strike a balance between attack effectiveness and stealthiness, we ultimately selected a poisoning rate of 10\% on BadSR.

\begin{figure*}[ht]
    \centering
    \includegraphics[width=\linewidth]{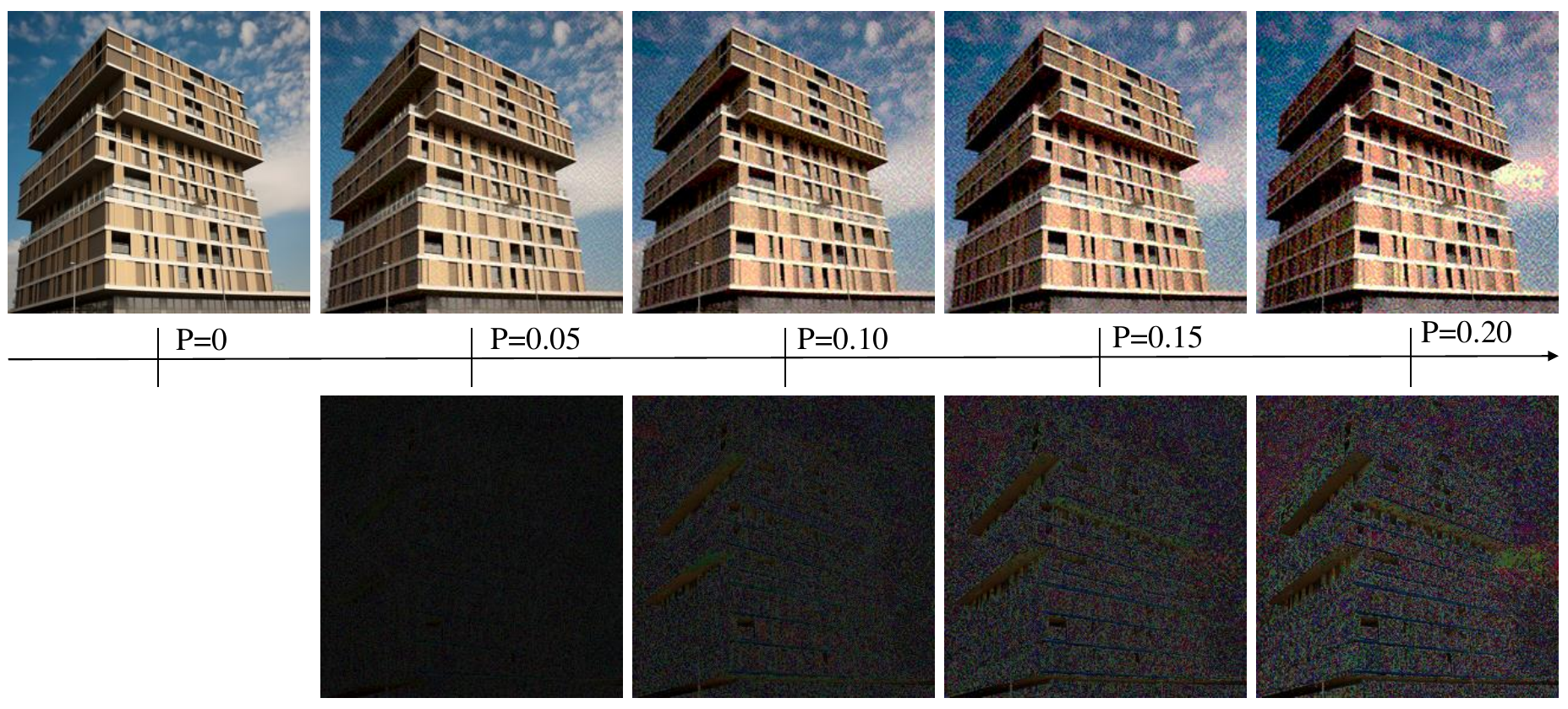}
    \caption{Visualization results of different perturbation budgets.}
    \label{fig:figure11}
\end{figure*}

\textbf{Perturbation budget.} The perturbation budget controls the maximum allowable perturbation in the image to generate HR label images. To investigate its effect on both attack effectiveness and stealthiness, we tested several values for the perturbation budget, ranging from 0.05 to 0.2. Our experiments show that lower perturbation budgets can still achieve a relatively good ASR, as shown in Table~\ref{tab:perturbation_budget}. Larger budgets increased ASR but also made the perturbation more detectable. After testing different values, We present in Figure~\ref{fig:figure11} the effects of applying different perturbation budgets to high-resolution images, along with visualizations of the differences compared to the original high-resolution images. A perturbation budget of 0.05 provides the best stealthiness. We ultimately chose a perturbation budget of 0.05 for generating the poisoned HR images, as it achieves a high ASR while maintaining good stealthiness.

\begin{table}[ht]
    \centering
    \caption{Impact of Perturbation Budget in BADSR. We use the backdoor LIIF model for evaluation. PSNR and SSIM are computed between the original HR images and the poisoned ones generated with different perturbation budgets. ASR indicates the attack success rate of the backdoored model under each budget.}
    \label{tab:perturbation_budget}
    \begin{tabular}{c|cccc}
    \toprule
    Method & 0.05 & 0.1 & 0.15 & 0.2 \\
    \midrule
    PSNR & 28.97 & 24.12 & 21.53 & 19.83 \\
    SSIM & 0.6895 & 0.4988 & 0.3995 & 0.3407 \\
    \midrule
    ASR & 85.73 & 87.82 & 88.56 & 88.93 \\
    \bottomrule
    \end{tabular}
\end{table}

\subsection{Computational Overhead}
\begin{figure}[ht]
    \centering
    \begin{subfigure}[b]{0.24\textwidth}
        \centering
        \includegraphics[width=\textwidth]{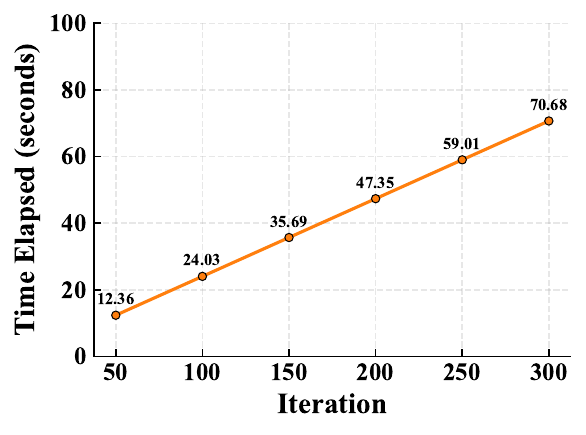}
        \caption{Time elapsed for generating poisoned LR}
    \end{subfigure}
    \hfill
    \begin{subfigure}[b]{0.24\textwidth}
        \centering
        \includegraphics[width=\textwidth]{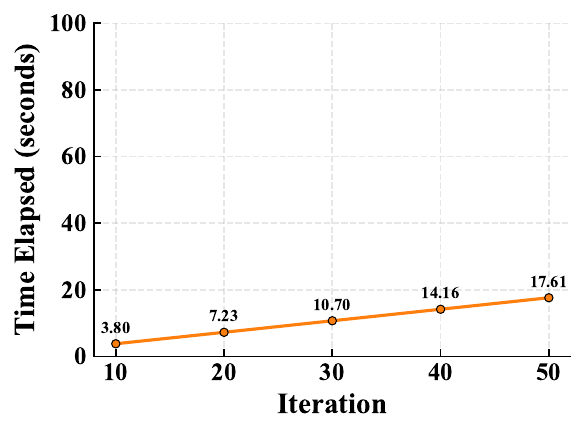}
        \caption{Time elapsed for generating poisoned HR}
    \end{subfigure}
    \caption{Computational overhead of BadSR.}
    \label{fig:Computational Overhead}
\end{figure}

Figure~\ref{fig:Computational Overhead} illustrates the computational overhead of generating a single poisoned LR and HR image on an Nvidia A800 GPU. The generation of a single LR image takes 70.67 seconds, whereas generating an HR image takes 17.61 seconds. This suggests that despite the increased complexity of our method, it still maintains an acceptable computational overhead, which is crucial to understanding its efficiency in practical applications.  Based on the convergence behavior of the loss during the generation process, we chose to generate the final poisoned LR image after 300 iterations, while the final poisoned HR image was generated after 50 iterations.

\begin{table}[ht]
    \centering
    \caption{Ablation study of effective poisoning.}
    \label{tab:ablation_study}
    \begin{tabular}{c|c|c|c}
    \toprule
        Poisoned Rate (\%) & Effective Poisoning & ASR & SSIM \\
        \midrule
        \multirow{2}{*}{5\%} & w/ & 68.42 & 0.8185 \\
             & w/o & 50.35 & 0.8534 \\
        \midrule
        \multirow{2}{*}{10\%} & w/ & 85.73 & 0.8122 \\
            & w/o & 80.10 & 0.8234 \\
        \midrule
        \multirow{2}{*}{20\%} & w/ & 91.02 & 0.7994 \\
            & w/o & 88.45 & 0.8087 \\
        \midrule
        \multirow{2}{*}{30\%} & w/ & 91.92 & 0.7642 \\
            & w/o & 90.22 & 0.7715 \\
    \bottomrule
    \end{tabular}
\end{table}

\subsection{Ablation Study}
In this section, we conduct an ablation study to assess the impact of effective poisoning. Effective poisoning selects poisoned samples based on the gradient of their contribution to the backdoor attack, using a genetic algorithm to choose the most impactful samples to enhance the backdoor effect. In contrast, without effective poisoning, the poisoned samples are randomly selected.

We train a backdoor LIIF model using the DIV2K dataset to assess the effect of effective poisoning. The results in Table~\ref{tab:ablation_study} show that although effective poisoning introduces slight perturbations to the generation of clean images, the model trained with effective poisoning significantly improves the ASR compared to the model without it. This shows that effective poisoning plays a crucial role in enhancing the effectiveness of backdoor attacks.

\subsection{Impact of Downstream Tasks}

\begin{figure}[ht]
    \centering
    \begin{subfigure}{\linewidth}
        \centering
        \includegraphics[width=\linewidth]{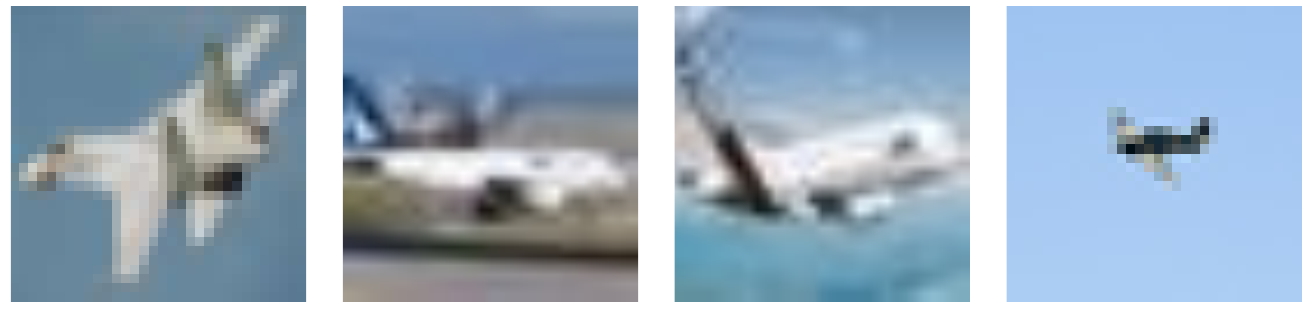}
        \caption{Origain image form CIFAR-10}
        \label{fig:figure13-1}
    \end{subfigure}
    \begin{subfigure}{\linewidth}
        \centering
        \includegraphics[width=\linewidth]{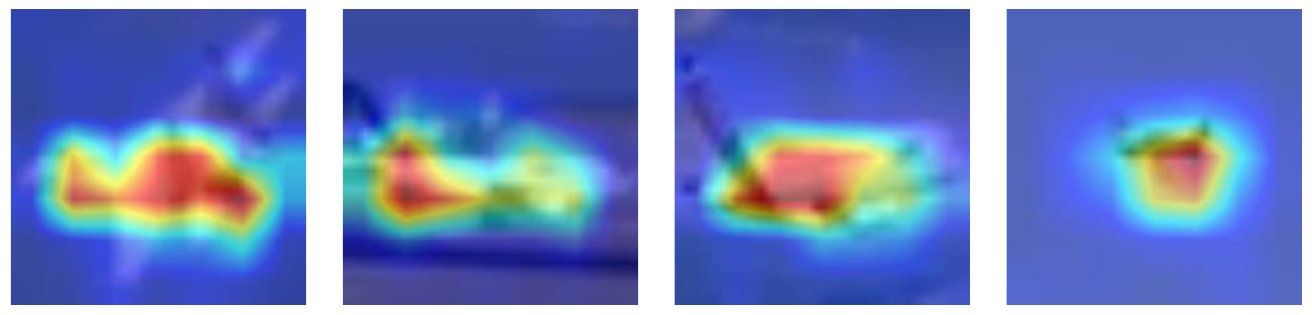}
        \caption{Grad-CAM of clean image}
        \label{fig:figure13-2}
    \end{subfigure}   
        \begin{subfigure}{\linewidth}
        \centering
        \includegraphics[width=\linewidth]{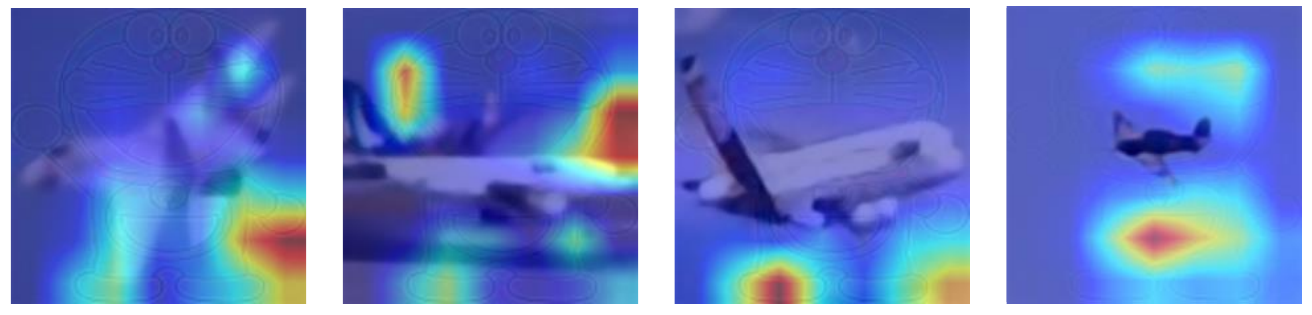}
        \caption{Grad-CAM of BadSR output image}
        \label{fig:figure13-3}
    \end{subfigure}  
    \caption{Grad-CAM of BadSR impact on image classification.}
    \label{fig:figure13}
\end{figure}

\begin{table}[ht]
    \centering
    \caption{Impact of image classification.}
    \label{tab:sr_classification_asr}
    \resizebox{\linewidth}{!}{
    \begin{tabular}{cc|cc|c}
    \toprule
        \multirow{2}{*}{\makecell{Upstream\\SR model}} 
        & \multirow{2}{*}{\makecell{Downstream\\classification\\model}} 
        & \multicolumn{2}{c|}{\makecell{Accuracy (\%)}} 
        & \makecell{ASR(\%)} \\
    \cmidrule{3-5}
        && \makecell{Clean model\\Clean img} & \makecell{Backdoor model\\Clean img} & \makecell{Backdoor model\\Backdoor img} \\
    \midrule
        \multirow{3}{*}{EDSR} & ResNet-50 & 91.23  & 89.44  & 78.64  \\
        & Vit-B  & 90.40  & 88.23  & 75.73  \\
        & MobileNet v2 & 90.53  & 89.14  & 76.36  \\
    \midrule
        \multirow{3}{*}{RCAN} & ResNet-50 & 89.50  & 86.90  & 69.80  \\
        & Vit-B  & 88.73  & 85.31  & 66.92  \\
        & MobileNet v2 & 88.92  & 86.00  & 68.13  \\
    \midrule
        \multirow{3}{*}{ESRGAN} & ResNet-50  & 88.62  & 84.55  & 71.27  \\   
        & Vit-B  & 87.73  & 83.40  & 68.34  \\
        & MobileNet v2 & 88.00  & 84.22  & 69.45  \\
    \midrule
        \multirow{3}{*}{SwinIR} & ResNet-50  & 90.03  & 87.92  & 73.61  \\
        & Vit-B  & 89.50  & 86.71  & 71.28  \\
        & MobileNet v2 & 89.76  & 87.34  & 72.02  \\
    \midrule
        \multirow{3}{*}{LIIF} & ResNet-50  & 89.81  & 87.25  & 74.90  \\
        & Vit-B  & 89.04  & 86.60  & 72.11  \\
        & MobileNet v2 & 89.50  & 86.90  & 73.20  \\
    \bottomrule
    \end{tabular}
    }
\end{table}

\begin{table}[ht]
    \centering
    \caption{Impact of object detection.}
    \label{tab:sr_objectdetection_asr}
    \resizebox{\linewidth}{!}{
    \begin{tabular}{cc|cc|c}
    \toprule
        \multirow{2}{*}{\makecell{Upstream\\SR model}} 
        & \multirow{2}{*}{\makecell{Downstream\\detection\\model}} 
        & \multicolumn{2}{c|}{\makecell{mAP(\%)}} 
        & \makecell{ASR(\%)} \\
        \cmidrule{3-5}
        && \makecell{Clean SR\\Clean img} & \makecell{Backdoor SR\\Clean img} & \makecell{Backdoor SR\\Backdoor img} \\
    \midrule
        \multirow{3}{*}{EDSR} 
        & MobileNetv2-YOLOv3 & 73.21 & 71.42 & 64.37 \\
        & Darknet53-YOLOv3   & 77.88 & 76.10 & 68.54 \\
        & EfficientNet-YOLOv3 & 75.34 & 73.98 & 66.73 \\
    \midrule
        \multirow{3}{*}{RCAN} 
        & MobileNetv2-YOLOv3 & 72.03 & 70.02 & 63.85 \\
        & Darknet53-YOLOv3   & 76.74 & 75.00 & 67.20 \\
        & EfficientNet-YOLOv3 & 74.80 & 73.05 & 65.32 \\
    \midrule
        \multirow{3}{*}{ESRGAN} 
        & MobileNetv2-YOLOv3 & 71.12 & 69.30 & 62.44 \\
        & Darknet53-YOLOv3   & 75.65 & 74.08 & 66.80 \\
        & EfficientNet-YOLOv3 & 73.42 & 71.76 & 64.19 \\
    \midrule
        \multirow{3}{*}{SwinIR} 
        & MobileNetv2-YOLOv3 & 74.45 & 72.68 & 65.90 \\
        & Darknet53-YOLOv3   & 78.21 & 76.55 & 70.21 \\
        & EfficientNet-YOLOv3 & 76.02 & 74.33 & 68.17 \\
    \midrule
        \multirow{3}{*}{LIIF} 
        & MobileNetv2-YOLOv3 & 73.66 & 71.60 & 64.80 \\
        & Darknet53-YOLOv3   & 77.12 & 75.30 & 69.45 \\
        & EfficientNet-YOLOv3 & 75.10 & 73.42 & 66.83 \\
    \bottomrule
    \end{tabular}
    }
\end{table}

To further evaluate the effectiveness of BadSR, we analyze its impact on two downstream tasks: image classification and object detection. We use officially pre-trained models and test them with images generated by the backdoored model. If the model misclassifies or fails to correctly detect objects, the attack is considered successful.

\subsubsection{Image Classification}
For image classification, we evaluate the impact of BadSR on three widely used image classification models: ResNet-50~\cite{resnet}, ViT-B~\cite{vit}, and MobileNet v2~\cite{mobilenetv2} in CIFAR-10~\cite{CIFAR-10}. As shown in Table~\ref{tab:sr_classification_asr}, the backdoor model achieves high ASR and ACC in all three models, indicating that the images generated by the BadSR-infected model can significantly impact downstream image classification models. To further illustrate this impact, we use Grad-CAM to visualize the differences between clean images and those generated by the backdoored model. As shown in Figure~\ref{fig:figure13}, the images produced by the backdoor model can mislead the classification model, resulting in incorrect predictions.   

\subsubsection{Object Detection}
We evaluated the impact of BadSR on downstream object detection models using three different detection architectures (MobileNetv2-YOLOv3~\cite{mobilenetv2,yolov3}, Darknet53-YOLOv3~\cite{yolov3}, and EfficientNet-YOLOv3~\cite{efficientnet,yolov3}) on Pascal VOC~\cite{pascal-voc-2012}. As shown in Table~\ref{tab:sr_objectdetection_asr}, BadSR-reconstructed clean images achieve a similar performance in downstream tasks compared to clean models, while also achieving a high ASR. This shows that BadSR can also be highly effective against downstream object detection models.

\section*{CONCLUSIONS}
This work explores the feasibility of implementing stealthy backdoor attacks in image super-resolution tasks. Specifically, we propose BadSR, a backdoor attack method designed for image super-resolution tasks. BadSR generates stealthily poisoned low-resolution images as triggers and corresponding poisoned high-resolution images as labels. To further enhance the effectiveness of the backdoor attack, we leverage the backdoor gradient to select efficient poisoned samples for the final poisoning data. In this way, the attacker can more stealthily implement a backdoor attack targeting image super-resolution models. Additionally, we investigate the impact of BadSR on downstream tasks. When applying a super-resolution model attacked by BadSR to enhance datasets for downstream tasks, anomalies appear in the downstream task performance. Extensive experiments demonstrate that BadSR is both stealthy and effective, and exhibits strong robustness. We hope this work will further contribute to the research on backdoor attacks in image super-resolution and spark increased attention to stealthy attacks.

\ifCLASSOPTIONcaptionsoff
  \newpage
\fi

\bibliographystyle{ieeetr}  
\bibliography{references}  


\end{document}